\documentclass{svproc}
\usepackage{float}
\usepackage{subfigure}
\usepackage{array}
\usepackage[a4paper, margin=1in]{geometry}
\usepackage{graphicx}
\usepackage{amsmath}
\usepackage[misc]{ifsym}
\usepackage[symbol]{footmisc}

\def\eg{\emph{e.g.,}}

\def\ie{\emph{i.e.,}}

\newcommand{\methodName}{SeFM}
\usepackage{url}

\begin{document}
\mainmatter              
\title{SeFM: A Sequential Feature Point Matching Algorithm for Object 3D Reconstruction}
\titlerunning{SeFM}  
%
\author{Zhihao Fang\inst{1} \and He Ma\inst{1}(\Letter) \and Xuemin Zhu\inst{1} \and Xutao Guo\inst{1} \and Ruixin Zhou\inst{2}}
\authorrunning{Zhihao Fang et al.} 
%
\tocauthor{Zhihao Fang, He Ma, Xuemin Zhu, Xuemin Zhu, Xutao Guo and Ruixin Zhou}
\institute{Sino-Dutch Biomedical and Information Engineering School, Northeastern University, Shenyang, China,\\
\email{mahe@bmie.neu.edu.cn}
\and
Zhiyuan College, Shanghai Jiao Tong University, Shanghai, China}

\maketitle              

\begin{abstract}
3D reconstruction is a fundamental issue in many applications and the feature point matching problem is a key step while reconstructing target objects. Conventional algorithms can only find a small number of feature points from two images which is quite insufficient for reconstruction. To overcome this problem, we propose \methodName\, a sequential feature point matching algorithm. We first utilize the epipolar geometry to find the epipole of each image. Rotating along the epipole, we generate a set of the epipolar lines and reserve those intersecting with the input image. Next, a rough matching phase, followed by a dense matching phase, is applied to find the matching dot-pairs using dynamic programming. Furthermore, we also remove wrong matching dot-pairs by calculating the validity. Experimental results illustrate that \methodName\ can achieve around $1,000$ to $10,000$ times matching dot-pairs, depending on individual image, compared to conventional algorithms and the object reconstruction with only two images is semantically visible. Moreover, it outperforms conventional algorithms, such as SIFT and SURF, regarding precision and recall.
\end{abstract}

\section{Introduction}
\label{sec:intro}
During the recent decades, 3D reconstruction is one of the key technologies used in many promising fields such as robot vision navigation, computer graphics based 3D games, video animation, Internet virtual roaming, e-commerce, digital library, visual communication, virtual reality, and \emph{etc}. In general, the 3D model is built based on a group of images captured from different angles and scales. One challenging task hence lies in estimating the spatial relationship between different images by matching feature points and achieving seamless reconstruction results. Many existing researches have made outstanding contributions to this field, \eg SIFT~\cite{Lowe04}, SURF~\cite{Herbert06} and PCA-SIFT~\cite{Yan04}. However, they can only utilize a small percentage of the information from the images which is not efficient. Moreover in conventional methods, even though two images can be mathematically matched by dot-pairs, this pairing relationship is rough and error-prone.

In order to overcome the above problem, this paper addresses a novel algorithm for feature point matching between different images named as~\methodName. The feature points are firstly obtained from images using conventional SIFT and SURF algorithm. Instead of searching feature points randomly and blindly, we search for the feature points by calculating the intersection between epipolar line of the camera and the edge of the targeted object. Thus, a pair of sequence of feature points are derived from each image, which we define as the \emph{rough matching phase}. Due to the different scale of the paired images, a linear interpolation algorithm is applied to help to find more matching feature points and this is defined as the \emph{dense matching phase}. Since there exist quantity wrong matchings due to noise and interpolating, it is possible to compute the maximum and minimum intensity of each point to retain high levels of reliability.

The main contribution of~\methodName~is its capability of generating the 3D model of a special target with a fairly small dataset of images, while conventional algorithms require using a large number of images from various angles. \methodName~increases the amount of information replenishment and increases the number of pixels available from each image by three orders of magnitude.

The rest of this paper is organized as follows: Section~\ref{sec:2} investigates the related literature. Section~\ref{sec:3} briefly introduces concepts and theorems of epipolar geometry in stereo vision as preliminary. Section~\ref{sec:4} presents the principle of point matching in detail. Section~\ref{sec:5} demonstrates the experimental results. Section~\ref{sec:6} concludes the whole paper.

\section{Related Research}\label{sec:2}
The 3D reconstruction problem based on Marr's visual theory framework forms a variety of theoretical methods. For example, according to the number of cameras, it can be classified into different categories such as monocular vision, binocular vision, trinocular vision, or multi-view vision~\cite{Lobay06,Wei12,Crandall13}; Based on the principle, it can be divided into regional-based visual methods and feature-based visual methods, model-based methods and rule-based visual methods, and \emph{etc.}~\cite{Ladicky17}; According to the way on obtaining data, it can be divided into active visual methods and passive visual methods~\cite{Crandall13,Ladicky17,Vu12}.

The reconstruction problem can mainly be divided into two folders as volumetric approaches and surface-based approaches. Volumetric approaches are usually used in reconstruction for medical data, while surface-based approaches are widely used for object reconstruction. One classic application is to build 3D city models from millions of images~\cite{Agarwal2011Building}. Another typical method is the structure-from-motion (SFM for short) algorithm~\cite{Koenderink1991Affine} and its variations. They all generated the 3D point cloud by matching feature points, calculating camera's fundamental matrix and essential matrix, and processing bundle adjustment.

One key step in 3D reconstruction is to find sufficient matching feature points and there exists a large amount of research on this topic~\cite{Donoser06,Moreels07}. At present, two conventional methods are designed based on the characteristics of corner and region respectively. The earliest corner detection algorithm was Moravecs corner descriptor~\cite{Moravec81}. Another well known approach was Harris corner detector~\cite{Harris88}, and it is simple and accurate, but sensitive to scale variation. This problem was overcame by Schmid and Mohr~\cite{Schmid97} using the Harris corner detector to identify interest points, and then creating a local image descriptor of them. The other category is the region detection algorithm. The most typical and widely used algorithms such as SIFT~\cite{Lowe04,Lowe99} and SURF~\cite{Herbert06,Bay08} fell into this category. The SIFT detectors have good robustness to image scale, illumination, rotation and noise. SIFT has plenty number of optimized variations on improving the efficiency including PCA-SIFT~\cite{Yan04}, Affine-SIFT (ASIFT)~\cite{Morel09} and \emph{etc}. Other methods aiming at improving the effectiveness and efficiency employ semantic scene segmentation~\cite{Kobyshev14}, local context~\cite{Wilson2014Network}.

All the above mentioned methods can be used to reconstruct the 3D model of a target object. However, to the best of our knowledge, the percentage of information utilized is low and this drives the reconstruction procedure requiring plenty number of images. This paper proposes a novel feature extraction algorithm with sequence matching and it can  can achieve a satisfactory reconstruction result with a small database.

\section{Review of Epipolar Geometry}\label{sec:3}
The preliminary of~\methodName~is to use epipolar geometry to identify the spatial relationship when we have two photos in different angles and positions. Epipolar geometry is the geometry of stereo vision between two views. It depends on the cameras' fundamental matrix, essential matrix and relative pose, and has no relation with the scene structure~\cite{Hartley00}.

\begin{figure}
	\subfigure[]{
		\includegraphics[width=0.48\textwidth]{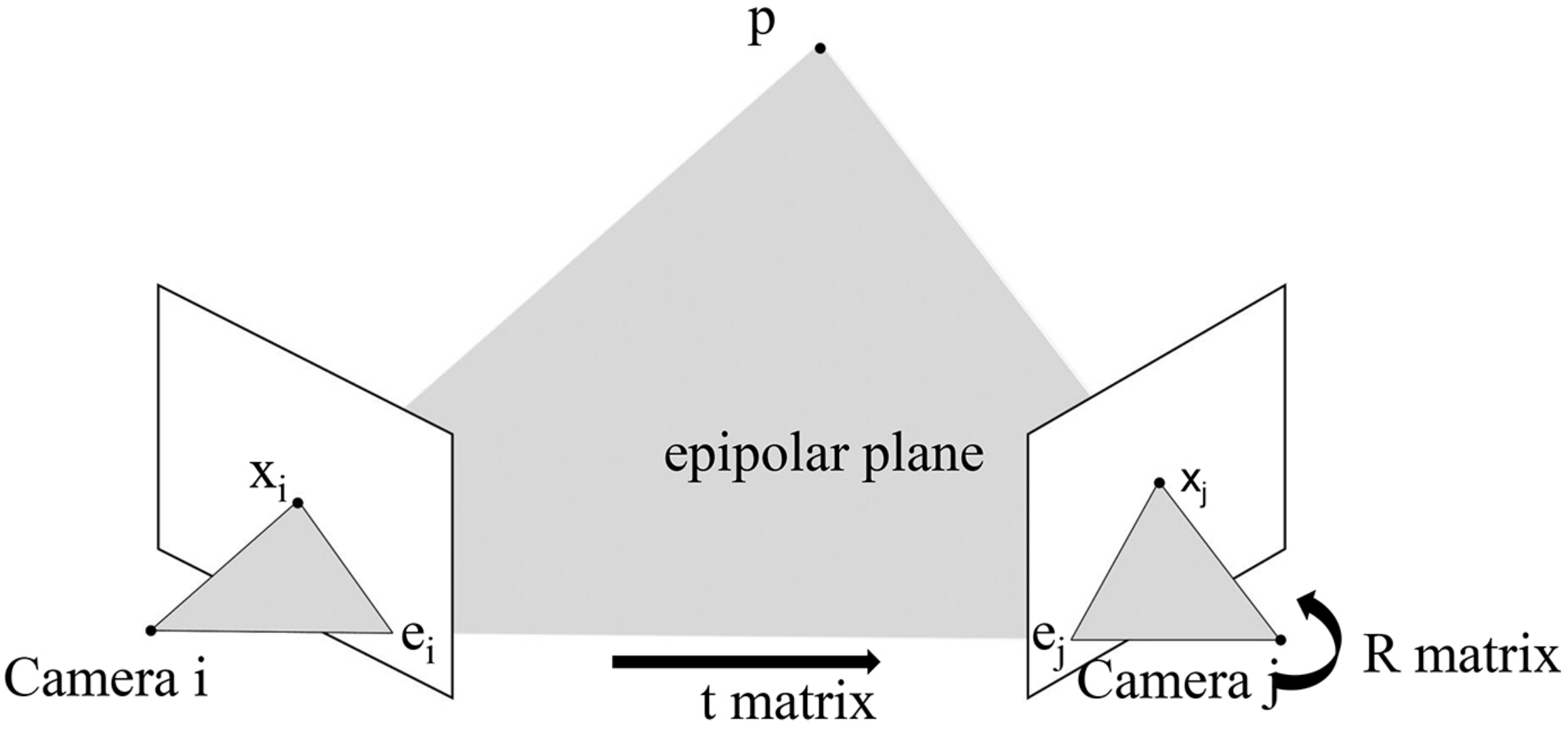}
		\label{figure:1a}
	}
	\subfigure[]{
		\hspace{2.4mm}
		\includegraphics[width=0.42\textwidth]{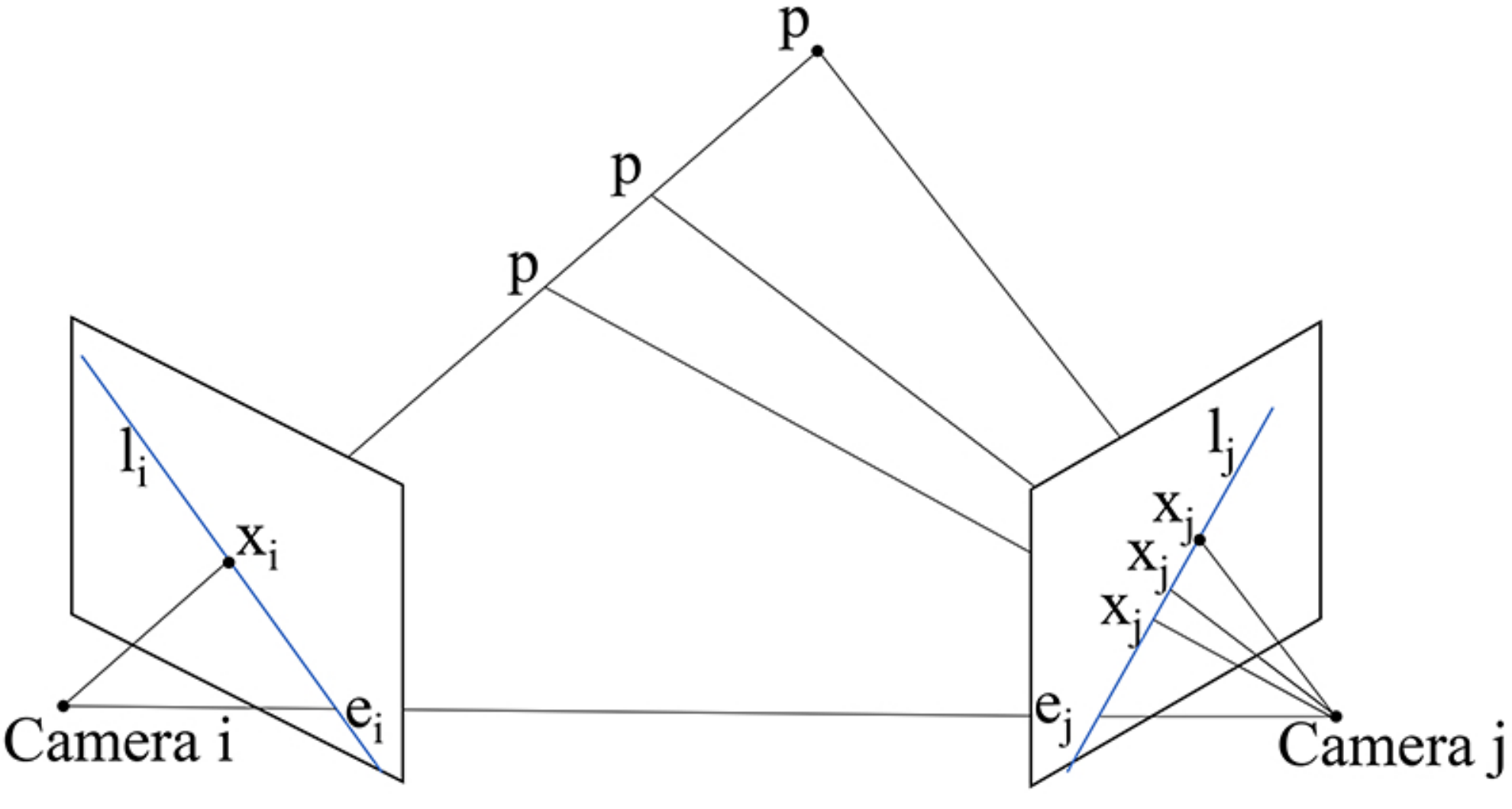}\\
		\label{figure:1b}
	}
	\caption{
		Illustration of views by two cameras: (a) implies the location relationship between Camera \emph{i} and \emph{j}. (b) shows that possible points \emph{p} in world coordinate determined by \emph{x$_i$} correspond to a series of \emph{x$_j$} in Line \emph{j}.}
\end{figure}

As shown in Figure~\ref{figure:1a}, the point \emph{p} taken by two different cameras from various locations, may appear to different positions in each image, in which \emph{X$_i$} and \emph{x$_i$} denotes the homogeneous and inhomogeneous coordinate in the \emph{i$^{\rm th}$} camera's view respectively. Note that the epipole is the point that is the intersection of line between camera locations and each image plane, denoted as $e_i$ and $e_j$ respectively. The location of $x_j$ can be calculated from that of $X_i$ as
\begin{equation}
\centering{} x_j=f\Pi(R_{ij}(X_i-t_{ij}))\text{,}
\end{equation}
where $\Pi$ is a function which converts homogeneous coordinate into inhomogeneous coordinate. \emph{t$_{ij}$} is the translation matrix, and \emph{R$_{ij}$} is the rotation matrix from camera \emph{i} to camera \emph{j}. Note that \emph{f} is relative to the internal reference of the camera.

In order to describe the location relationship between cameras in the world coordinate, the fundamental matrix (\emph{F}) should be used (\ie a 3$\times$3 matrix), where $F$ contains the information of translation, rotation and the intrinsics of both cameras. To present calibrated views in epipolar geometry, Formula~(\ref{formula:2})
\begin{equation}
\centering{} X_j^\top FX_i=0
\label{formula:2}
\end{equation}
can be extended to Formula~(\ref{formula:3}) from points to lines:
\begin{equation}
\centering{} l_j^\top FX_i=0\text{.}
\label{formula:3}
\end{equation}

As Figure~\ref{figure:1b} shows, Formula~(\ref{formula:2}) and~(\ref{formula:3}) reveal that a certain location \emph{x$_i$} of point \emph{p} can derive \emph{x$_j$} must be in a certain line \emph{l$_j$} in the corresponding coordinate and vice versa. However, even \emph{x$_i$} is known, the precise coordinate \emph{x$_j$} is difficult to be ascertained by geometric computation, while arbitrary line \emph{l$_i$} and its corresponding line \emph{l$_j$} in the other view can be confirmed. Since $F$ has seven degrees of freedom, the sufficient conditions to get unique solution is obtaining at least eight matching dot-pairs of images~\cite{Longuet-Higgins81}. Combining some information of the scene, the conditions even can be reduced to five matching dot-pairs~\cite{Nister04}.


\section{\methodName\ Algorithm}\label{sec:4}
\begin{figure}[ht]
	\centering
	\includegraphics[width=0.8\textwidth]{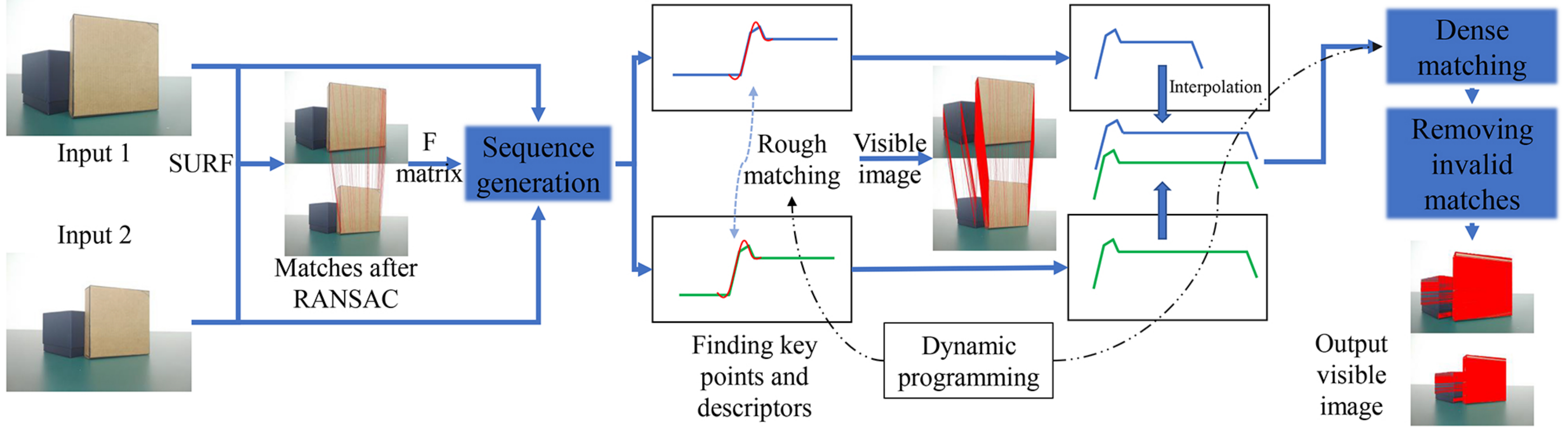}
	\caption{The whole procedure of using \methodName\ to search for matching dot-pairs.}
	\label{figure:AD}
\end{figure}

For simplicity, the input images are all taken by calibrated cameras in the algorithm. The basic idea of~\methodName\ is to find a maximum number of matching dot-pairs between different images and the most challenging part is to solve scale-variant problem. Figure~\ref{figure:AD} illustrates the whole procedure of \methodName. The SURF algorithm followed by random sample consensus algorithm ($RANSAC$ for short) is first applied to find a few matching feature points. We then calculate the fundamental matrix $F$ and generate the sequence of points. Afterwards, the rough matching phase and dense matching phase are then processed using dynamic programming. We hence obtain the matching points between two images. Finally, all the matching points are achieved after removing invalid matches. Each step is detailed as follows.

\subsection{Sequence Generation}
SIFT, as a point descriptor, possesses the scale-invariant feature~\cite{Lowe04}. SURF, based on SIFT but faster than it, is another approach to search dot-pairs~\cite{Herbert06}. Using K-d tree, the processing time of SURF on feature points matching can be reduced to one second. As stated in Section~\ref{sec:3}, it is possible to calculate of $F$ and $E$ since sufficient dot-pairs can be easily obtained using SURF. Although SURF has higher accuracy than most existing matching algorithms, it is still necessary to adopt $RANSAC$ to get rid of the wrong matchings.

\begin{figure}[!h]
	\centering
	\begin{minipage}[t]{0.5\textwidth}
		\centering
		\includegraphics[width=6cm]{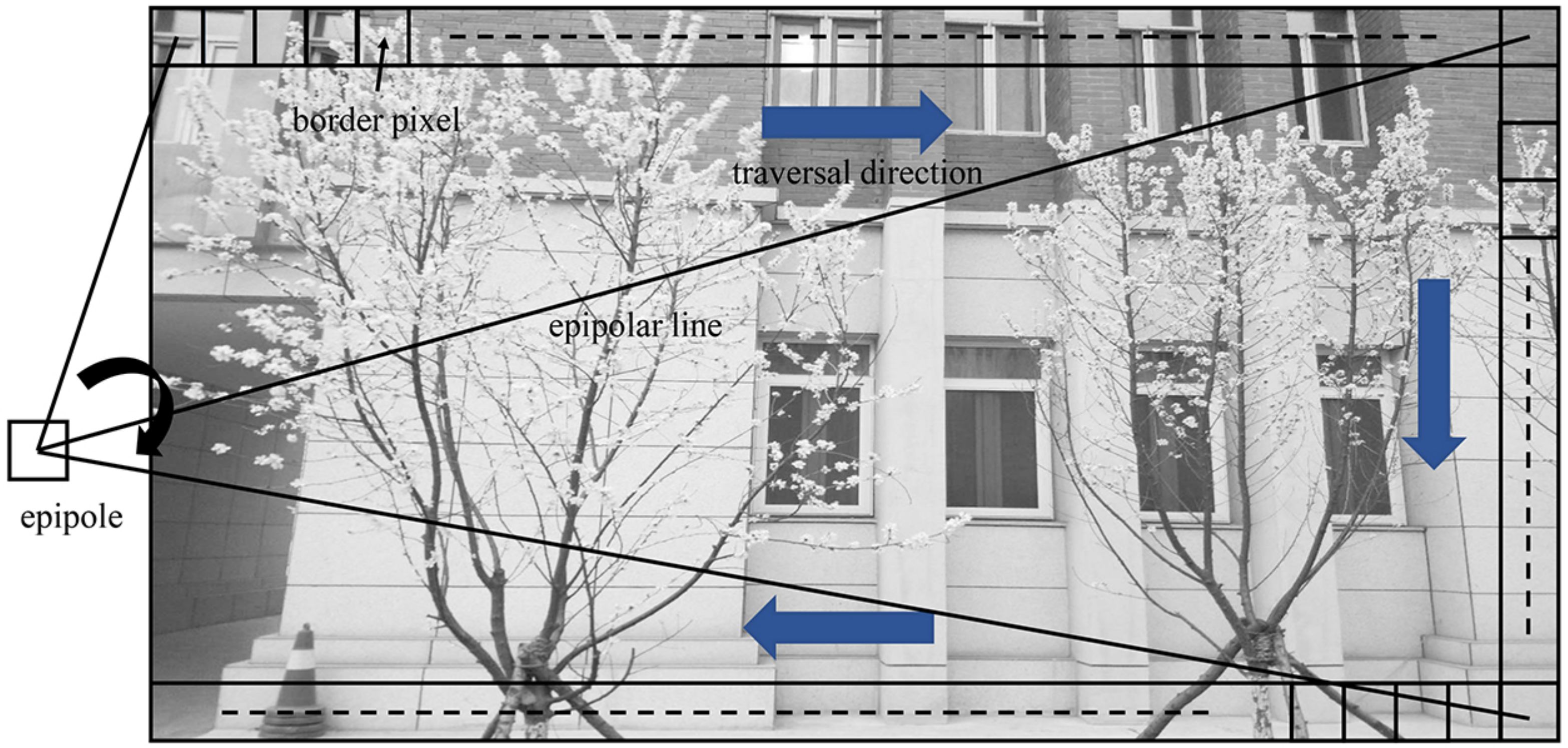}
		\caption{The traversal scan of epipolar lines. For example, the epipole is in the left bottom corner area. Epipolar line sequences are those lines intersect with the captured image.}
		\label{figure:2}
	\end{minipage}
	\hspace*{10pt}
	\begin{minipage}[t]{0.45\textwidth}
		\centering
		\includegraphics[width=4cm]{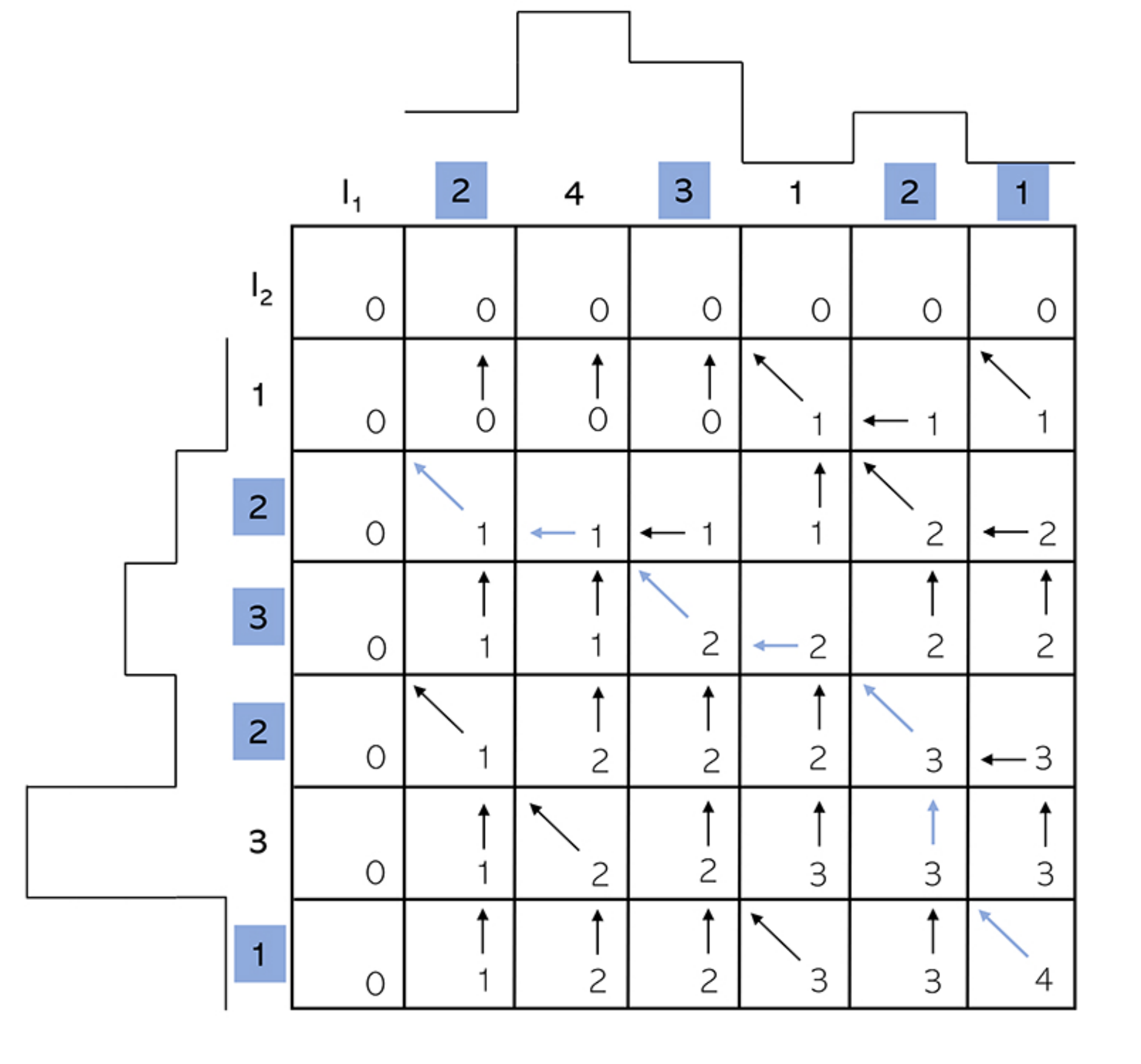}
		\caption{The sequence of line \emph{l$_1$} is above the graph and the sequence of line \emph{l$_2$} is on the left side, where numbers are referred to intensity. Arrows based on scores imply the process of tracing back.}
		\label{figure:3}
	\end{minipage}
\end{figure}

As shown in Formula~(\ref{formula:3}), the corresponding relationship between the epipolar lines ($CREL$ for short) from two images can be achieved. Note that the line $l_i$ is bound to pass through the epipole. In order to achieve the maximum number of epipolar lines, a traversal scan with epipole as center is applied to obtain every pair of $CREL$ sequences (as shown in Figure~\ref{figure:2}).

Suppose there exist two images $\Pi_1$, and $\Pi_2$, the epipolar lines can be represented with a set of points as
\begin{equation}
l_{1}=\{p_{11}, p_{12},\cdots\ ,p_{1i},\cdots\},\
l_{2}=\{p_{21}, p_{22},\cdots\ ,p_{2i},\cdots\}\text{,}\label{formula:6}
\end{equation}
where \emph{l$_1$} is the set of epipolar lines of $\Pi_1$, and \emph{l$_2$} is the corresponding line satisfying Formula~(\ref{formula:3}). \emph{p$_{ni}$} ($n=1,2;\;i=1,2,\cdots$) are consistent sequential points in the line \emph{l$_n$}.

\subsection{Dynamic Programming}\label{sec:4_2}
As epipolar lines are sequential, searching the optimal solution of the point matches between the two epipolar lines from different cameras can be deemed as longest common sequence ($LCS$ for short) problem. Dynamic programming ($DP$ for short), a basic method in algorithm field, is suitable for solving these problems by establishing recurrence~\cite{Ohta85}. Once the recursive expressions is obtained, optimal solution will be available by tracing back (shown in Figure~\ref{figure:3}). For push-broom multiple views, semi-global matching~\cite{Heiko08} has a good performance. However, in other circumstances, the relationship between pixels can only be conformed through epipolar line. In this way, \methodName\ employs the naive dynamic programming algorithm to find the edge of the target object.
\subsection{The Rough Matching Phase}\label{sec:4_3}
It is tricky to match points in $CREL$ directly when camera locations are distant. Though the scales of images vary from view to view, the edge of the target object is distinct and easy to find due to the intensity changing significantly. Therefore, we first try to find the edge of target object using $DP$ method stated in Section~\ref{sec:4_2} along a certain epipolar line and treat the two intersection as the ending points of a section. We define this process as \emph{the rough matching phase}. Define the change of intensity of a certain point as
\begin{equation}
\mathrm{d}a_i=\left|\frac{a_{i-1}-a_{i+1}}{2}\right|\text{,}
\end{equation}
where \emph{a$_i$} is the intensity of the point \emph{p$_i$}. Combining with Formula~(\ref{formula:6}), it can be expressed as
\begin{equation}
a_1=\left|l_1\otimes\alpha\right|,\ a_2=\left|l_2\otimes\alpha\right|,\
\alpha=\begin{pmatrix}
\dfrac{1}{2} & 0 & -\dfrac{1}{2}
\end{pmatrix}\text{.}
\end{equation}

\begin{figure}[H]
	\centering
	\includegraphics[width=12cm]{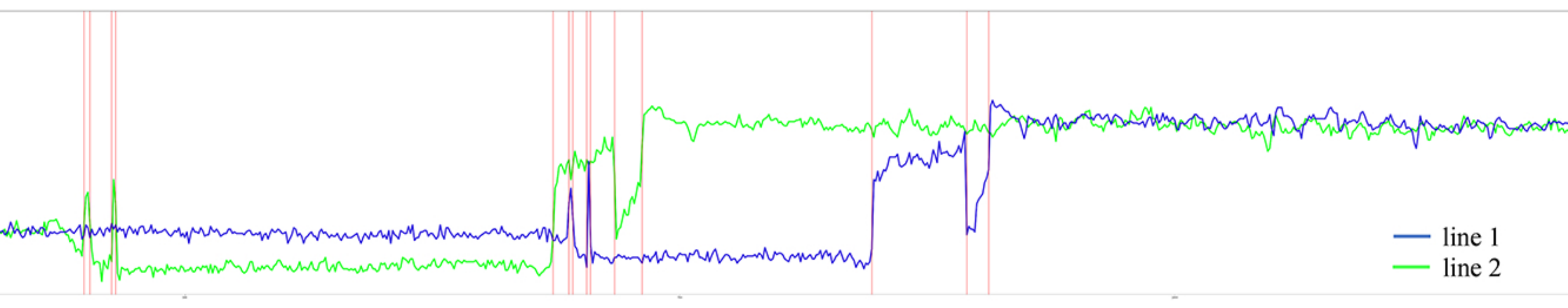}
	\caption{Line graph of the epipolar line-pair and rough matching key points. The x-axis presents the point along an epipolar line and the y-axis presents the intensity. Points in the two lines are referred to intensity from points in $CREL$ sequence. The intensity jumps are locations of key points in rough matching.}
	\label{figure:4}
\end{figure}

Figure~\ref{figure:4} is a part of line graph describing the intensity of points on \emph{$l_1$} (in blue) and \emph{$l_2$} (in green). Positions marked with red lines are points whose $\mathrm{d}a$ are larger than the threshold, which are considered as key points of the target object.
Afterwards, the rough-matching point sets
\begin{equation}
RM_1=\{P_{11},P_{12},\cdots,P_{1i},\cdots\},\ RM_2=\{P_{21},P_{22},\cdots,P_{2i},\cdots\}
\end{equation}
are obtained, where \emph{P$_{1i}$}, \emph{P$_{2i}$} are key points of \emph{l$_1$} and \emph{l$_2$} in rough matching. Note that \emph{RM$_1$} and \emph{RM$_2$} are subsets of \emph{l$_1$} and \emph{l$_2$}.

To match the points in \emph{RM}, appropriate descriptors describing them take on extreme importance. In this process, not only the intensity of key point itself, but that of the points around it in $l_n$ should be considered. We recommend using Fourier expansion of the adjacent points to describe \emph{P$_{1i}$} and \emph{P$_{2i}$}. The Fourier expansion is
\begin{equation}
\begin{split}
f(x)&=\sum_{n=1}^{\infty}(a_n\cos n\omega x+b_n\sin n\omega x +a_0)\text{,}\\
a_0=\frac{1}{2\pi}\int_{-\pi}^{\pi}f(x)\mathrm{d}x\text{,}\
a_n&=\frac{1}{\pi}\int_{-\pi}^{\pi}\cos nxf(x)\mathrm{d}x\text{,}\
b_n=\frac{1}{\pi}\int_{-\pi}^{\pi}\sin nxf(x)\mathrm{d}x\text{,}
\end{split}
\end{equation}
where \emph{a$_n$} and \emph{b$_n$} provide a good description of the overall adjacent points. Considering the discreteness of points and computational efficiency of the algorithm, a fast Fourier transformation (FFT) is applied.

\subsection{Dense Matching}\label{sec:4_4}
\begin{figure}[ht]
	\centering
	\includegraphics[width=11cm]{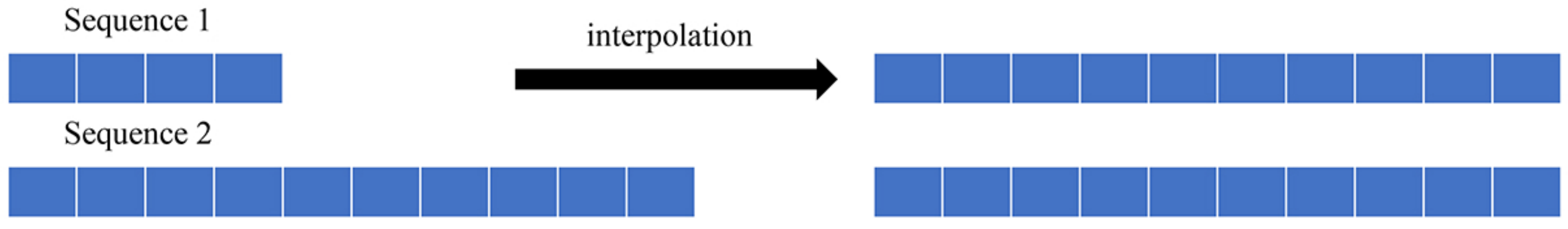}
	\caption{Illustration of sequence interpolating. Since the length of Sequence 1 is shorter, a linear interpolating is applied to Sequence 1 to make its length equal to Sequence 2.}
	\label{figure:5}
\end{figure}

Once the rough matching is completed, adjusting the scale of original sequence became practical. In order to establish the corresponding relationship between $RM_1$ and $RM_2$, the original sequence can be tailored into several subsequences. For example in Figure~\ref{figure:5}, Sequence 1 and 2 are from \emph{l$_1$} and \emph{l$_2$}. $P_{1begin}$ matches $P_{2begin}$ and $P_{1end}$ matches $P_{2end}$, where $P_{1begin}\in RM_1$, $P_{1end}\in RM_1$, $P_{2begin}\in RM_2$, and $P_{2end}\in RM_2$. To normalize the scale, a linear interpolating algorithm is used to equalize the length of Sequence 1 and 2.
\vspace*{-10pt}

In this part, since the matching procedure is straightforward, it leads to possible wrong matchings due to noise and the interpolating process. 
Defining $I^+$ and $I^-$ as
\begin{equation}
I^+ = \frac{I(x_i)+I(x_{i+1})}{2},\ I^- = \frac{I(x_i)+I(x_{i-1})}{2}\text{,}
\end{equation}
the maximum and minimum intensity of the points can be defined as
\begin{equation}
I_{max}=\max\{I^-,I(x_i),I^+\}, \ I_{min}=\min\{I^-,I(x_i),I^+\}\text{,}
\end{equation}
where \emph{I} is the intensity of a point, and \emph{x} is the point location of the subsequence after interpolating. In this way, the cost function can be written as
\begin{equation}
C(x_{i}, y_j)=\max\{0, I(y_j)-I_{min}, I_{max}-I(y_j)\}\text{,}
\end{equation}
which can effectively reduce the amount of computation and the influence of noise.

\subsection{Inconsistency Problem}\label{sec:4_5}
\begin{figure}[H]
	\setlength{\fboxsep}{0.2mm}
	\subfigure[]{
		\fbox{
			\includegraphics[width=0.48\textwidth]{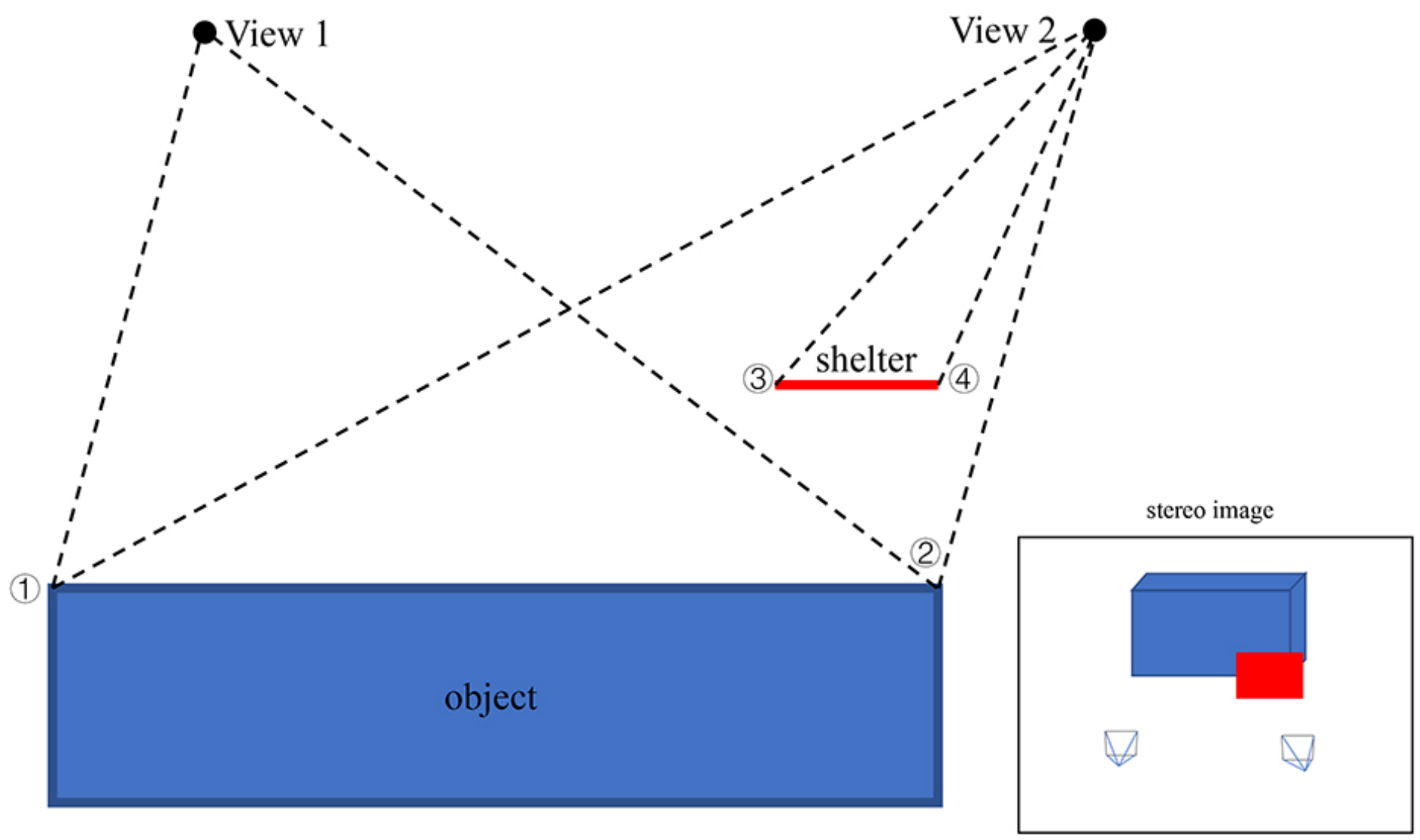}
			\label{figure:6a}}
	}
	\subfigure[]{
		\hspace{-2.0mm}
		\fbox{
			\includegraphics[width=0.48\textwidth]{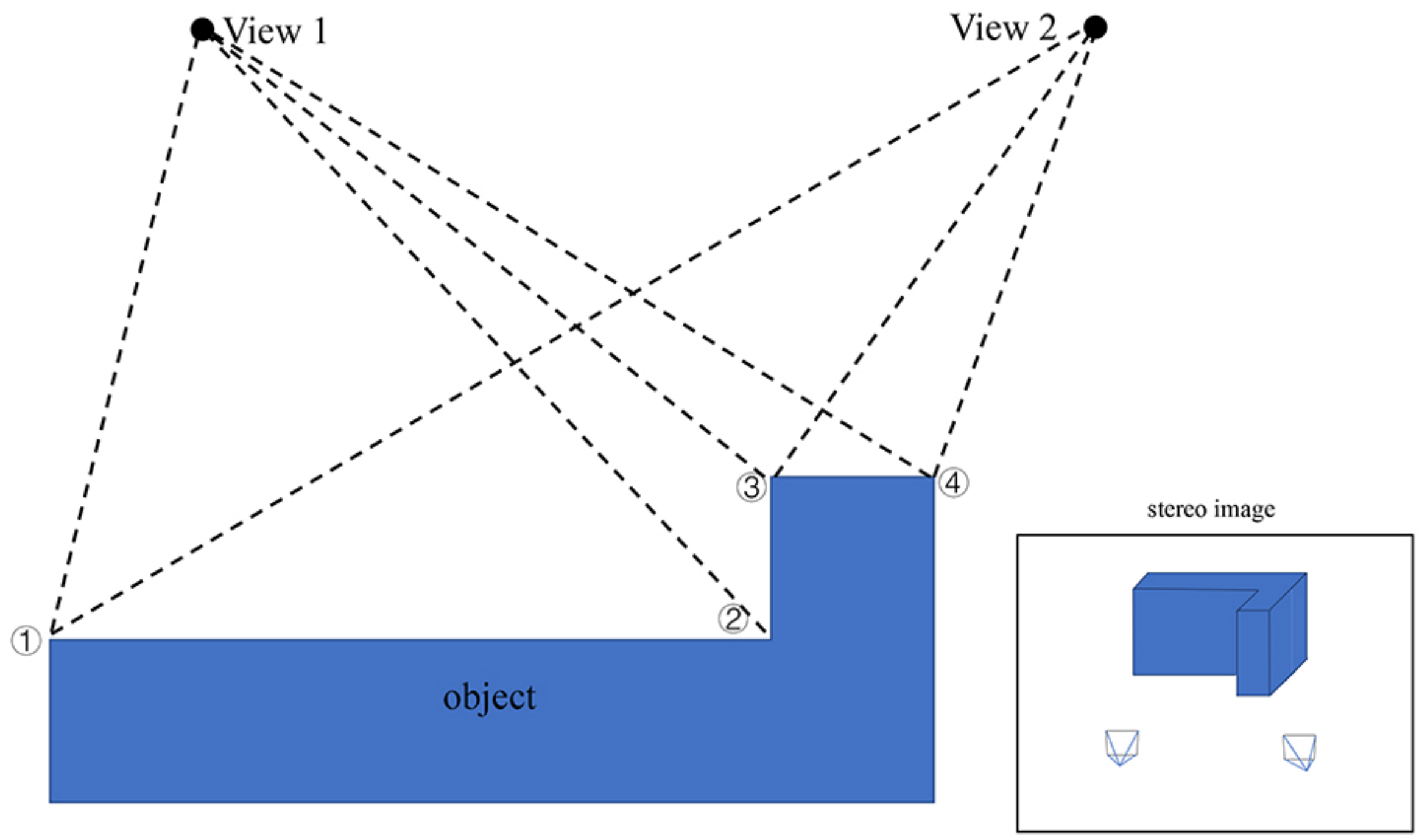}
			\label{figure:6b}}
	}
	\caption{Occlusion situation. (a) shows one situation when object is shaded by a shelter, in which View 2 can search two more key points \textcircled{3}\textcircled{4} than View 1. (b) shows another situation when a part of object is shaded due to its shape, where only View 1 could search position \textcircled{2} of two views.}
\end{figure}

In reality, the occlusion problem must be taken into consideration. \methodName\ handles this problem from imagewise to pixelwise matching level. As is shown in Figure~\ref{figure:6a}, due to the occlusion in front of the object, two redundant key points will be searched in View 2 during rough matching process. Another case shown in Figure~\ref{figure:6b}, because of the object's shape, one key point will vanish in View 2.

To solve this problem, the validity $S_v$ of the tailored sequences must be considered. If there exists no other $P$ from $P_{begin}$ to $P_{end}$, the dense sequence between them is valid, which can be concluded as
$$
S_v=
\begin{cases}
\mathrm{valid} &\mathrm{if}\ P_{end}\ is\ \mathrm{next\ to}\ P_{begin}\ \mathrm{in\ }RM\\
\mathrm{invalid} &\mathrm{otherwise}
\end{cases}
$$

From the perspective of pixelwise matching, the depth information, which can be calculated from the dot-pairs, matrix $R$ and $t$, may contribute to the validation of matches. Given the depth of each points \emph{D$_{pi}$}, the validity of each point \emph{S$_{pi}$} can be determined from
$$
S_p=
\begin{cases}
\mathrm{valid} &\mathrm{if}\ \|D_{p(i+1)}-D_{pi}\|\le \mathrm{threshold}\\
\mathrm{invalid} &\mathrm{otherwise}
\end{cases}
$$

\section{Experimental Results}\label{sec:5}
\begin{figure}
	\newcommand{\pwidth}{2cm}
	\newcommand{\width}{1.95cm}
	\begin{tabular}{p{1.5cm}<{\centering} p{\pwidth}<{\centering}p{\pwidth}<{\centering}p{\pwidth}<{\centering}p{\pwidth}<{\centering}p{\pwidth}<{\centering}}
		\footnotesize Index & a & b & c & d & e\\
		\footnotesize Input image 1 &
		\begin{minipage}{\width}
			\includegraphics[width=\width]{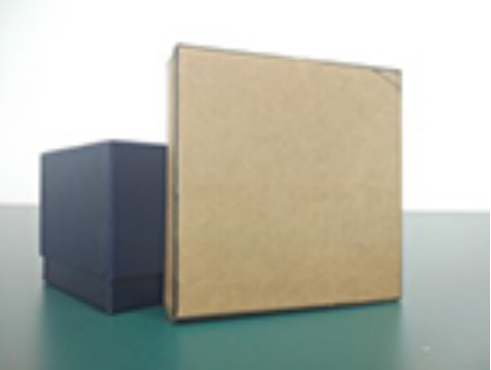}
		\end{minipage} &
		\begin{minipage}{\width}
			\includegraphics[width=\width]{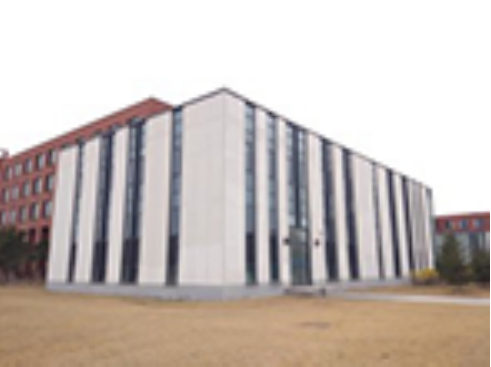}
		\end{minipage} &
		\begin{minipage}{\width}
			\includegraphics[width=\width]{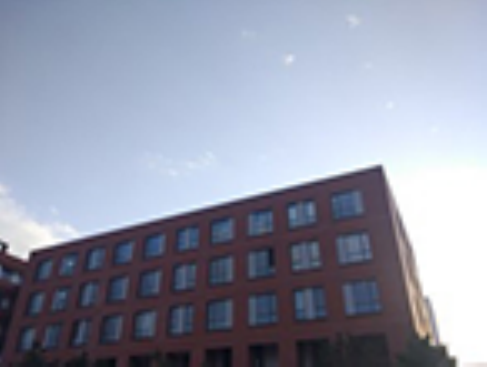}
		\end{minipage} &
		\begin{minipage}{\width}
			\includegraphics[width=\width]{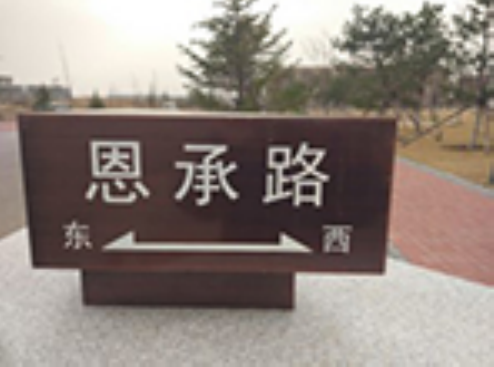}
		\end{minipage} &
		\begin{minipage}{\width}
			\includegraphics[width=\width]{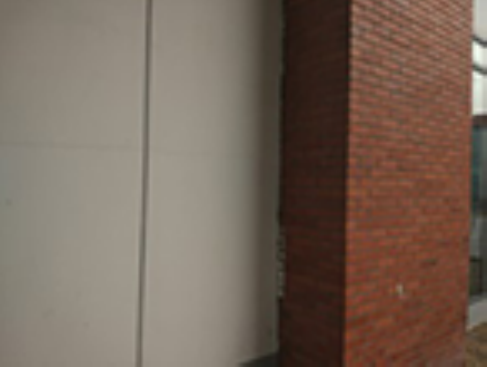}
		\end{minipage}\vspace{0.05cm}\\
		\footnotesize Input image 2 &
		\begin{minipage}{\width}
			\includegraphics[width=\width]{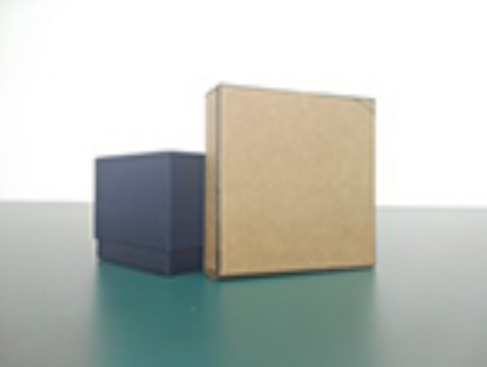}
		\end{minipage} &
		\begin{minipage}{\width}
			\includegraphics[width=\width]{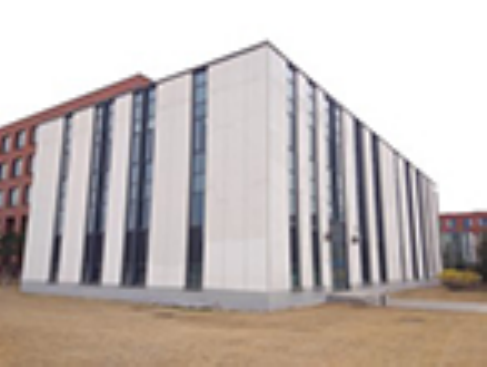}
		\end{minipage} &
		\begin{minipage}{\width}
			\includegraphics[width=\width]{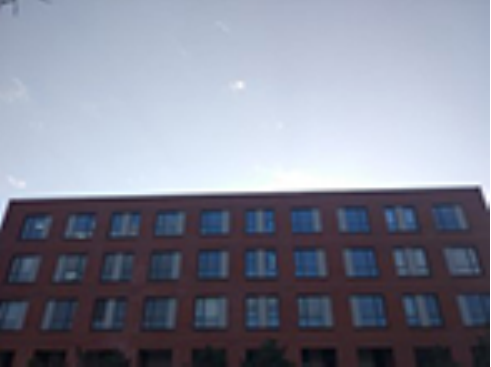}
		\end{minipage} &
		\begin{minipage}{\width}
			\includegraphics[width=\width]{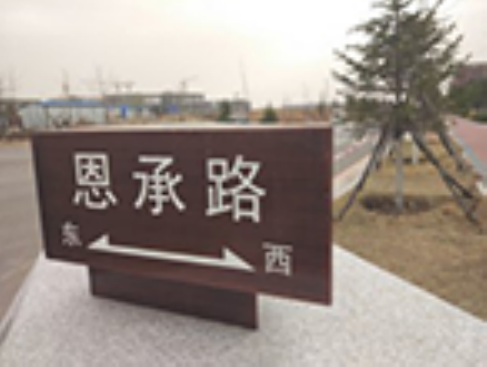}
		\end{minipage} &
		\begin{minipage}{\width}
			\includegraphics[width=\width]{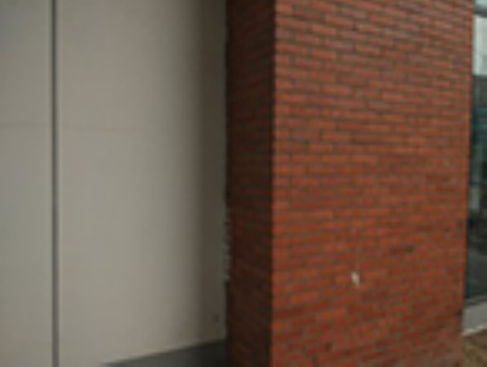}
		\end{minipage}\vspace{0.05cm}\\
		\vspace{-0.7cm} \footnotesize Matching by SIFT + SURF &
		\begin{minipage}{\width}
			\includegraphics[width=\width]{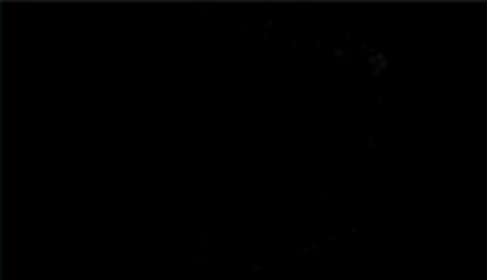}
		\end{minipage} &
		\begin{minipage}{\width}
			\includegraphics[width=\width]{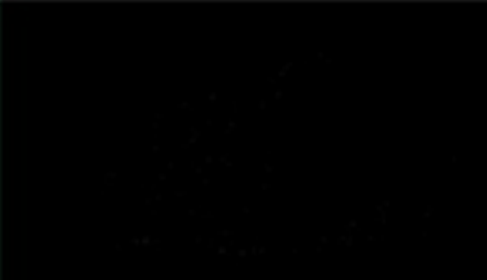}
		\end{minipage} &
		\begin{minipage}{\width}
			\includegraphics[width=\width]{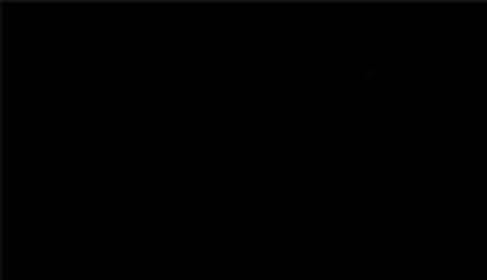}
		\end{minipage} &
		\begin{minipage}{\width}
			\includegraphics[width=\width]{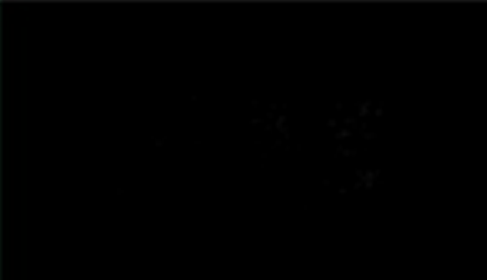}
		\end{minipage} &
		\begin{minipage}{\width}
			\includegraphics[width=\width]{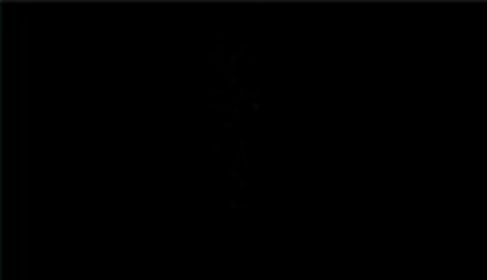}
		\end{minipage}\vspace{0.05cm}\\
		\vspace{-0.85cm} \footnotesize Matching by SIFT + SURF (amplification) &
		\begin{minipage}{\width}
			\includegraphics[width=\width]{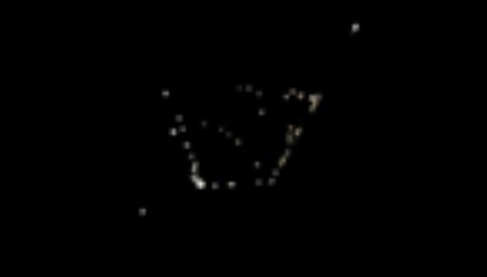}
		\end{minipage} &
		\begin{minipage}{\width}
			\includegraphics[width=\width]{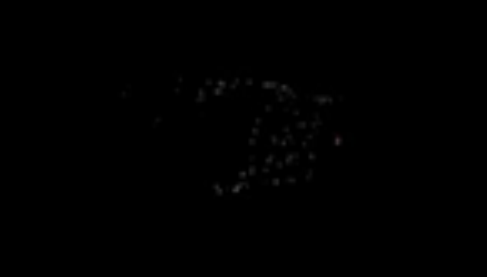}
		\end{minipage} &
		\begin{minipage}{\width}
			\includegraphics[width=\width]{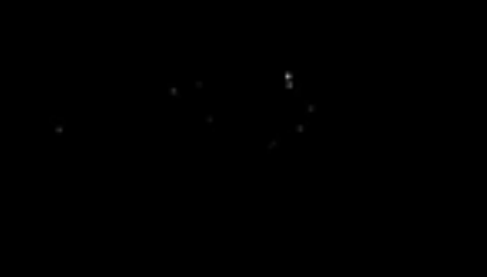}
		\end{minipage} &
		\begin{minipage}{\width}
			\includegraphics[width=\width]{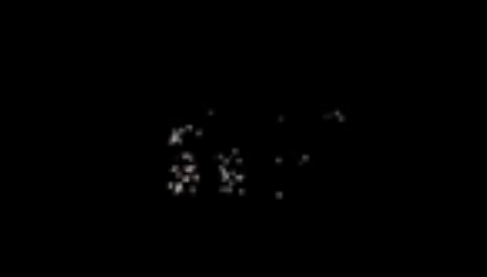}
		\end{minipage} &
		\begin{minipage}{\width}
			\includegraphics[width=\width]{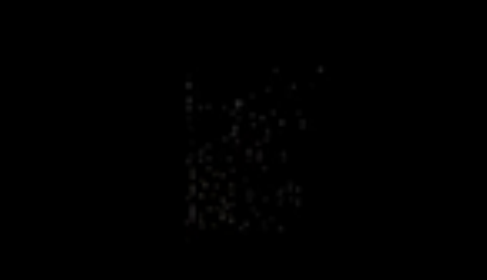}
		\end{minipage}\vspace{0.05cm}\\
		\ \footnotesize Matching by \methodName &
		\begin{minipage}{\width}
			\includegraphics[width=\width]{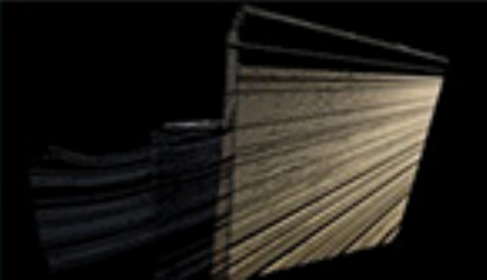}
		\end{minipage} &
		\begin{minipage}{\width}
			\includegraphics[width=\width]{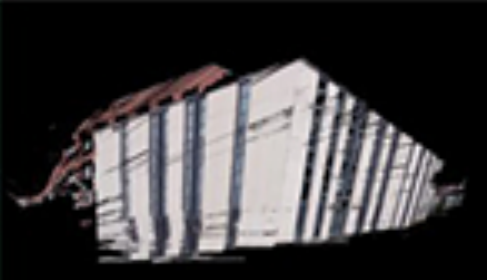}
		\end{minipage} &
		\begin{minipage}{\width}
			\includegraphics[width=\width]{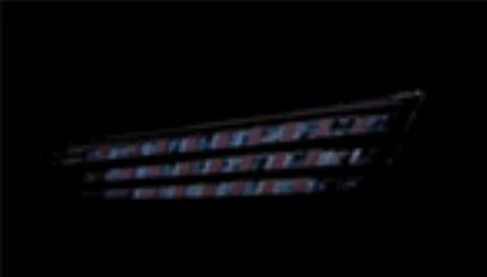}
		\end{minipage} &
		\begin{minipage}{\width}
			\includegraphics[width=\width]{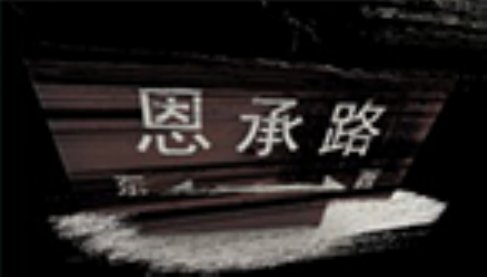}
		\end{minipage} &
		\begin{minipage}{\width}
			\includegraphics[width=\width]{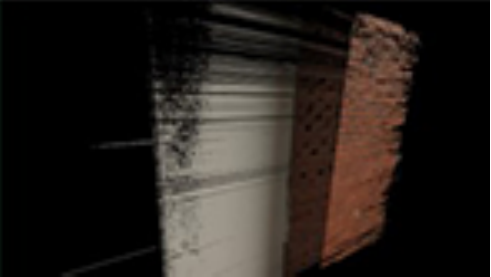}
		\end{minipage}\vspace{0.05cm}\\
	\end{tabular}
	\vspace{0.3cm}
	\caption{Exhibition of results from part of datasets. Row 1 and 2 are input images. The rest lines are visualization of point matching in 3D reconstruction. Row 3 and 4 are respective results of direct and amplified images using SIFT+SURF. Row 5 are results using \methodName .}
	\label{figure:7}
\end{figure}
The image dataset of our experiment came from various objects, each of which was taken in two directions by a Mi 5 smartphone with Sony IMX298 camera sensor. All images are originally with a resolution of 4608$\times$3456, and resized to 2000$\times$1500 considering of computational efficiency. Our experiments were conducted on one PC with 2.7 GHz Intel Core i7-7500U CPU, 16 GB memory, and the implementation with Python on points matching costs around 11 minutes.

\paragraph{Number of matches} In Figure~\ref{figure:7}, we exhibit some typical 3D reconstruction results without mesh using only two input images from the dataset. The selected images cover multiple cases, \eg~objects with different scales, objects with different backgrounds, objects with different textures. It is very difficult to reconstruct the object from two images by simply using SIFT and SURF, since the number of matching dot-pair is too small. Observed from Figure~\ref{figure:7}, the number of point clouds matched by SIFT and SURF is too small to be distinguished, even if we enlarge the figures in Row 3 to Row 4. On the contrary, \methodName\ can reconstruct the target object to a fairly good shape. \methodName\ outperforms compared with SIFT and SURF matching algorithm in the number of matching points, also listed in Table~\ref{table:1}. The reason lies in that \methodName\ can get such dense points with calculating the epipolar relationship between images in the process of matching.

\begin{table}[ht]
	\renewcommand\arraystretch{1}	
	\caption{The statistics of matching on \methodName\ and SIFT+SURF.}	\begin{tabular}[H]{p{2cm}<{\centering}p{2cm}<{\centering}p{2cm}<{\centering}p{2cm}<{\centering}p{2cm}<{\centering}}
		\hline
		input images index & number of matches from \methodName\ & number of matches from SIFT+SURF & precision & recall \\
		\hline
		a & 2,023,399 & 186 & 0.965063 & 0.8567428\\
		b & 249,782 & 13 & 0.997638 & 0.5983016 \\
		c & 1,058,315 & 171 & 0.968988 & 0.7091526 \\
		d & 725,602 & 311 & 0.826024 & 0.4216738 \\
		e & 140,264 & 114 & 0.999537 & 0.8300612 \\
		\hline
	\end{tabular}
	\vspace*{-12pt}
	\label{table:1}
\end{table}

\begin{figure}[!h]
	\centering
	\begin{minipage}[t]{0.35\textwidth}
		\centering
		\includegraphics[width=2.8cm]{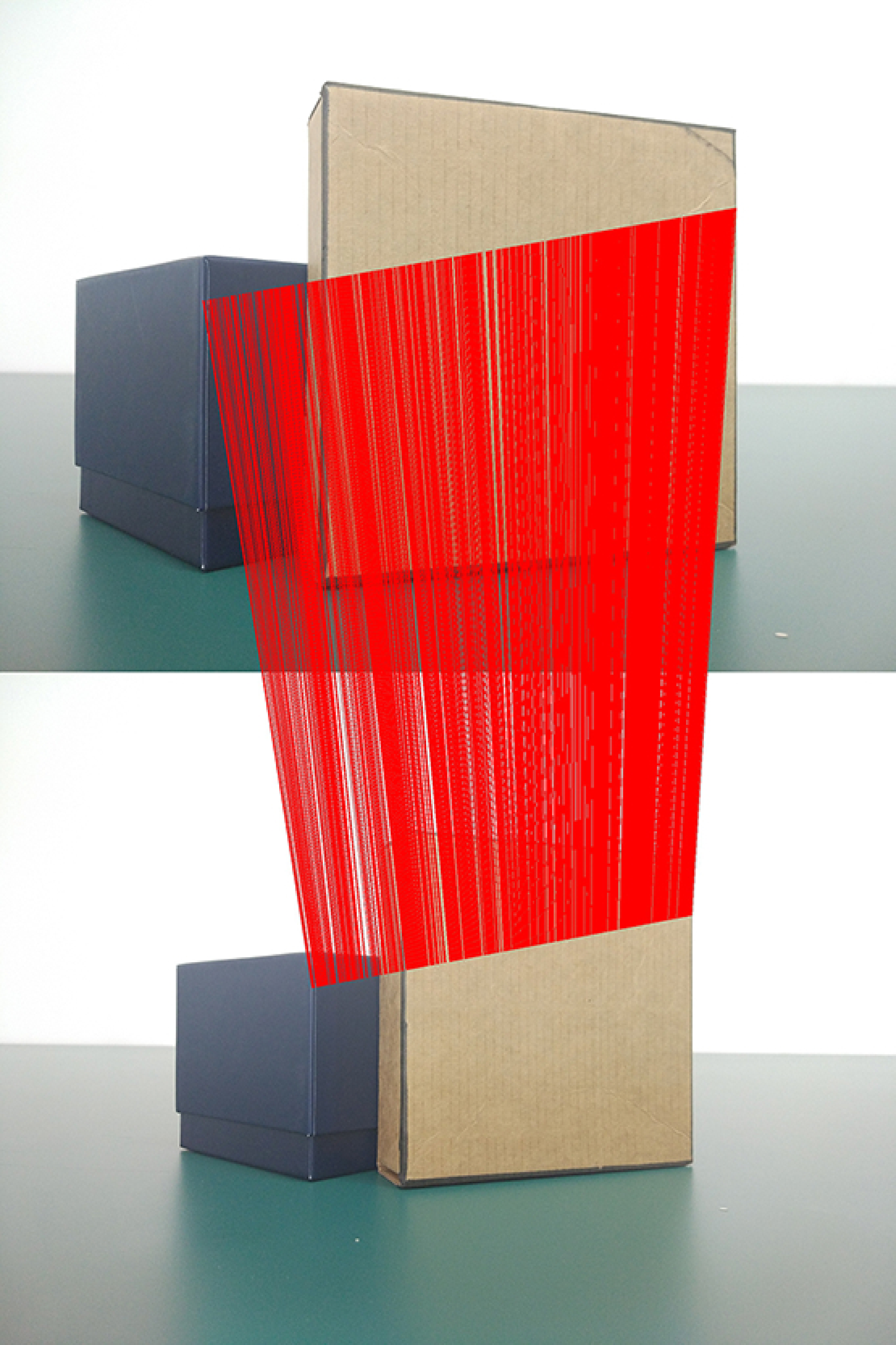}
		\caption{Point matching result where only one of epipolar line pairs is applied. }
		\label{figure:8}
	\end{minipage}
	\hspace*{10pt}
	\begin{minipage}[t]{0.6\textwidth}
		\centering
		\includegraphics[width=5.5cm]{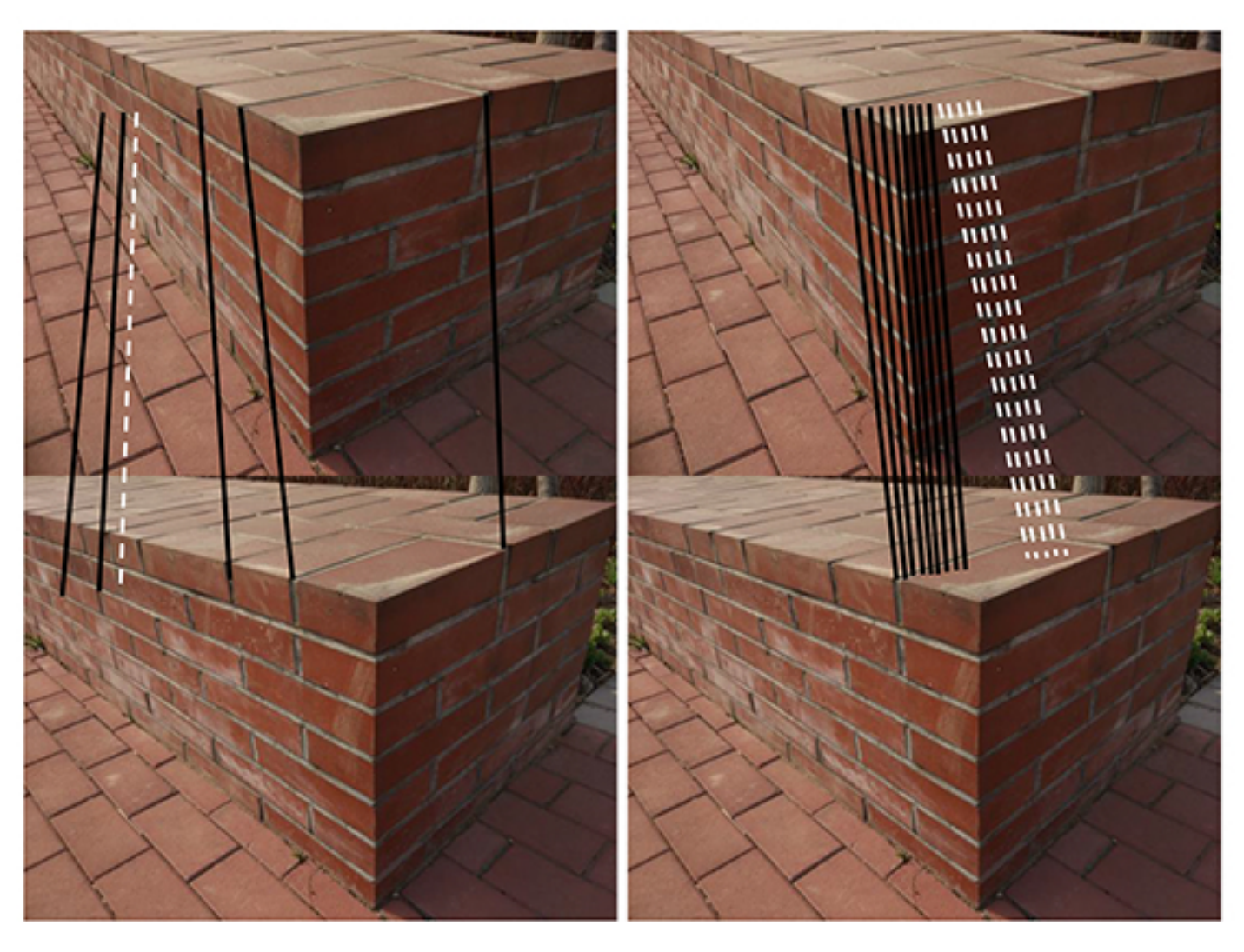}
		\caption{Examples of two wrong matching. The left one is situation of missing key points and the right one is dislocation-match.}
		\label{figure:9}
	\end{minipage}
\end{figure}

\paragraph{Precision and Recall} Precision and Recall are widely used in algorithm evaluation. In order to evaluate the accuracy of \methodName\, we use $precision$ and $recall$ defined as follows: \emph{\# of positives} stands for the number of key points in the input image. Given the inconformity of numbers and scales between two input images, we use the average number of both. \emph{\# of matches} records the total number of matching points between two images. \emph{\# of correct-positives} is the number of right matches where two points are exactly the same. Thus, \emph{precision} and \emph{recall} can be calculated as below:
$$
\begin{aligned}
precision=\frac{\#~of~correct\textnormal{-}positives}{\#~of~matches}, \
recall=\frac{\#~of~correct\textnormal{-}positives}{\#~of~positives}
\end{aligned}
$$
As presented in Table~\ref{table:1}, the input images (a) and (e) are simple, in which input (a) are two photos of two cuboids on the table and input (e) are two images of a wall. Thus, they achieve higher recall. Conversely, the input images (b) and (d) are more complexed on the object surface, which lead to the lower recall. In input image (d), there exist trees at the upper right of that image which are not main focus of the target, so the value of precision and recall is lower compared with other inputs. Consequentially, \methodName\ can achieve a high value of precision and a good value of recall. This proves that most of matching found by \methodName\ are correct. Moreover, it can find an average of $68.3\%$ of matching points between two input images, which are around $1,000$ to $10,000$ times compared to conventional SIFT+SURF algorithm, while the time consuming is around 500 times than that of SIFT+SURF due to sequential point matching. As Figure~\ref{figure:8} shows, even scales differ in two input images, the point matching result is still dense. There is no doubt that \methodName\ algorithm makes outstanding performances in matching as few as only two images.

\paragraph{Wrong matching Case Analysis}
In \methodName, the most probable wrong matches are due to key-point missing in the rough matching phase and the dislocation-match in the dense matching phase. As illustrated in Figure~\ref{figure:9}, the situation of missing key points tend to occur in objects with large sum of similar features, \eg~windows and doors. These objects are sensitive to the lighting when photographing. In fact, this kind of wrong matching hardly arises when images are taken virtually the same time in a day. The dislocation-match is a critical fault in the process, triggering the chain reaction of continual deterioration in dynamic programming. Somehow, $DP$ reduces the probability of such phenomenon happening.

\section{Conclusion}\label{sec:6}
This paper introduced a novel algorithm of points matching, which can utilize a maximum number of valid information from images. Based on the spatial relation of the images, it can reconstruct the object well with two or more random images. Compared to previous points matching, our approach can find a significant number of matching dot-pairs. It is proven that \methodName\ will be effective on 3D reconstruction, panoramic images fields and \emph{etc}.

%
%

\end{document}